\documentclass[lettersize,journal]{IEEEtran}
\usepackage{amsmath,amsfonts}
\usepackage{algorithmic}
\usepackage{algorithm}
\usepackage{array}
\usepackage[caption=false,font=normalsize,labelfont=sf,textfont=sf]{subfig}
\usepackage{textcomp}
\usepackage{stfloats}
\usepackage{url}
\usepackage{verbatim}
\usepackage{booktabs}
\usepackage{graphicx}
\usepackage{multirow}
\usepackage{utfsym}
\usepackage{svg}
\usepackage{amssymb}
\usepackage{cite}
\usepackage{makecell}
\begin{document}

\title{CLIP-FSAC++: Few-Shot Anomaly Classification with Anomaly Descriptor Based on CLIP}

\author{Zuo Zuo, Jiahao Dong, Yao Wu, Yanyun Qu~\IEEEmembership{Member, IEEE,} Zongze Wu~\IEEEmembership{Member, IEEE}

\thanks{Manuscript created November, 2024. (Corresponding author: Zongze Wu.)

Zuo Zuo is with National Key Laboratory of Human-Machine Hybrid Augmented Intelligence, National Engineering Research Center for Visual Information and Applications, and Institute of Artificial Intelligence and Robotics, Xi'an Jiaotong University, Xi’an 710049, China (e-mail: Nostalgiaz@stu.xjtu.edu.cn).

Jiahao Dong is with Guangdong Laboratory of Artificial Intelligence and Digital Economy (SZ) (e-mail: dongjiahao2023@email.szu.edu.cn).

 Yao Wu and Yanyun Qu are with the School of Informatics, Xiamen University, Xiamen 361005, China (e-mail: wuyao@stu.xmu.edu.cn, yyqu@xmu.edu.cn)

Zongze Wu is with College of Mechatronics and Control Engineering Shenzhen University, Shenzhen, China (e-mail: zzwu@szu.edu.cn).}}

\markboth{Journal of \LaTeX\ Class Files,~Vol.~14, No.~8, August~2021}%
{Shell \MakeLowercase{\textit{et al.}}: A Sample Article Using IEEEtran.cls for IEEE Journals}


\maketitle

\begin{abstract}
Industrial anomaly classification (AC) is an indispensable task in industrial manufacturing, which guarantees quality and safety of various product. To address the scarcity of data in industrial scenarios, lots of few-shot anomaly detection methods emerge recently. In this paper, we propose an effective few-shot anomaly classification (FSAC) framework with one-stage training, dubbed CLIP-FSAC++. Specifically, we introduce a cross-modality interaction module named Anomaly Descriptor following image and text encoders, which enhances the correlation of visual and text embeddings and adapts the representations of CLIP from pre-trained data to target data.  In anomaly descriptor, image-to-text cross-attention module is used to obtain image-specific text embeddings and text-to-image cross-attention module is used to obtain text-specific visual embeddings. Then these modality-specific embeddings are used to enhance original representations of CLIP for better matching ability. Comprehensive experiment results are provided for evaluating our method in few-normal shot anomaly classification on VisA and MVTEC-AD for 1, 2, 4 and 8-shot settings. The
source codes are at https://github.com/Jay-zzcoder/clip-fsac-pp.

\end{abstract}


\section{Introduction}
\IEEEPARstart{I}{ndustrial} anomaly classification (IAC)~\cite{zuo} included by industrial anomaly detection (IAD)~\cite{simplenet, ofc, tbs} is one of the most widely used computer vision technologies aiming to classify normal and abnormal samples in the course of manufacturing and production, which guarantees the quality and safety of industrial products. In industrial scenarios, rare appearance of anomalous product leads to difficulty in collecting abnormal samples. Meanwhile, data annotation is labor-intensive and time-consuming. Therefore, industrial anomaly detection methods regard anomaly classification as one-class classification~\cite{deepocc,ldf} and detect anomalies in an unsupervised manner to address scarcity of anomalous samples, that is, these methods only use normal samples for training without any annotations. However, unsupervised anomaly detection can not meet needs once and for all. In many cases, there are only a few normal samples in the beginning, which is called cold start~\cite{coftad}.

\begin{figure}
  \centering
  \centerline{\includegraphics[width=\linewidth]{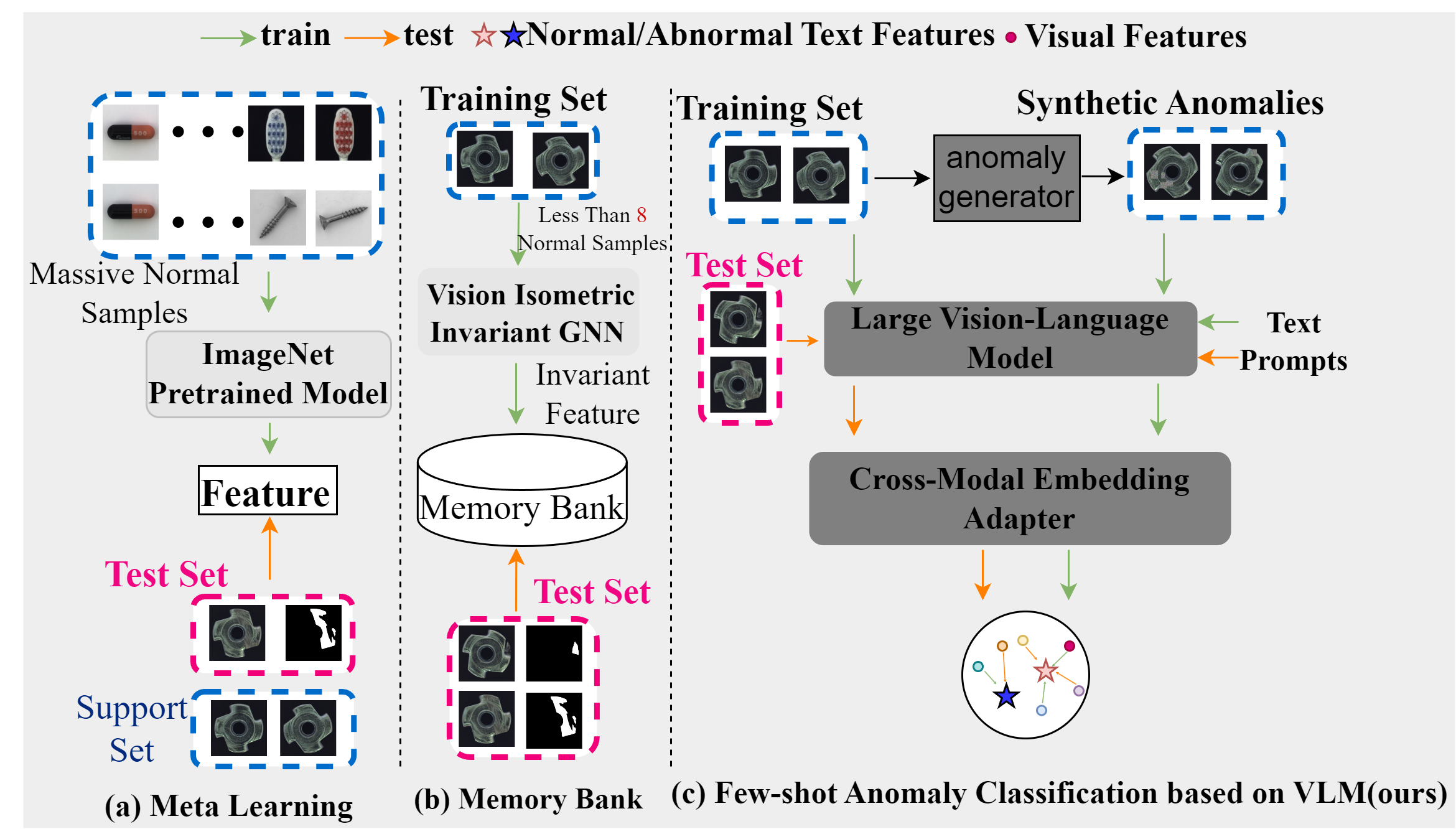}}
  \caption{Different diagrams for few-shot anomaly detection. (a) few-shot unsupervised AC in meta learning. (b) few-shot  unsupervised AC based on vision isometric invariant GNN using memory bank. Our proposed (c) leverages alignment capability between image and text of large vision-language model and fine-tunes it for few-shot AD without extra memory bank and massive normal samples.}
  \label{fewshot}
\end{figure}

To make a smooth transition facing with cold start, many few-shot anomaly detection (FSAD) methods~\cite{al, zsv} are proposed recently, which are trained on only a few normal samples. One prevailing paradigm is to solve FSAD by employing meta-learning (See Fig. \ref{fewshot} (a)). Since the difficulty in collecting massive similar normal images, Graphcore~\cite{graphcore} considers constructing a memory bank, which only needs a few normal images of target category (See Fig. \ref{fewshot} (b)). However, it neglects the burden brought by the additional inference time.

Pre-trained vision-language models (VLMs)~\cite{minigpt4, flamingo} have gained substantial attention recently. In FSAC, WinCLIP proposes a compositional prompt ensemble (CPE) on state words and prompt templates and aggregation of window/patch/image-level features aligned with texts based on CLIP framework. It performs well due to strong generalization ability. AnomalyCLIP~\cite{anomalyclip} proposes object-agnostic text prompts and a global abnormality loss function for zero-shot anomaly detection. The success of these methods is attributed a lot to strong  recognition and generalization ability of CLIP~\cite{clip}. As mentioned in CLIP-FSAC~\cite{clipfsac}, utilizing weight-pretrained encoders of CLIP as our training paradigm not only enables us to capture broader concepts with open vocabulary but also avoids introducing additional computational overheads brought by modules.

The performance of CLIP in FSAC drastically deteriorates owing to two crucial points. On the one hand, text prompts which describe normality and anomalies and captured visual-language relationship play an important role in FSAC. In our earlier exploration, we change different combination of texts in zero-shot anomaly classification with CLIP and find that results vary widely. It demonstrates that it is a tough nut to crack to find accurate prompts in anomaly classification. Inaccurate text prompts lead to vision-language mismatch. On the other hand, there exists a large distribution discrepancy between industrial and pre-trained natural images, which gives rise to insufficient visual representation. To this end, we propose a novel few-normal-shot anomaly classification framework, dubbed CLIP-FSAC++ based on CLIP-FSAC~\cite{clipfsac}, focusing on leveraging the complementary advantage of cross-modality fusion to solve FSAC task.

In details, to easily utilize strong representative ability of CLIP in anomaly classification,  we introduce light-weight image and text adapters~\cite{clipadapter} and anomaly descriptor following CLIP encoders to adapt prior representations. Due to absence of abnormal samples in training set,  we synthesize anomaly samples in the beginning. Then we optimize proposed modules via matching visual and text feature corresponding to normality and abnormality. Prior visual representations are transformed by image adapter into more discriminative space in which visual features can be grouped into normal and abnormal clusters. To detect anomaly samples, we use text features representing normality and abnormality to match visual features. As mentioned above, finding optimum text prompts is difficult. So we add text adapter to redistribute text features specializing for anomaly classification to bridge the prompts gap.  Besides, we propose a cross-modality interaction module named anomaly descriptor consisting of two cross-modality attention modules. Anomaly descriptor incorporates modality-specific features and makes vision and language representations compatible for each other. Compared to CLIP-FSAC, we simplify fine-tuning process by substituting two-stage training strategy with joint training strategy. In a word, the main contributions of this paper are summarized as four-fold:
\begin{itemize}
\item{We empirically analyze that the designing of text
prompts is important in anomaly classification
and propose CLIP-FSAC++, which fully explores the potential of CLIP in few-shot anomaly classification.}
\item{We propose to adapt CLIP for anomaly classification with less trainable parameters and less training samples. The distribution discrepancy can be alleviated and text features are optimized to target domain. CLIP-FSAC++ can be easily transferred into FSAC.}
\item{We introduce Anomaly Descriptor (AD) consisting of image-to-text cross-attention module and text-to-image cross-attention module. AD incorporates modality-specific prior and calculates cross-modality attention to improve visual and language representative ability for better vision-language correlation.}
\item{We have conducted thorough experiments. Experimental results show superior performance of CLIP-FSAC++ in few-shot anomaly classification. CLIP-FSAC++ outperforms previous few-shot methods on VisA and MVTEC-AD for 1-shot, 2-shot, 4-shot and 8-shot, even surpassing many full-shot anomaly detection methods.}
\end{itemize}
Compared with the conference version, we mainly expand the following contents for the present work:
\begin{itemize}
\item{We discard two-stage training strategy and jointly optimize all trainable modules, which simplifies training process and saves training cost. Compared to two-stage training strategy, simplified joint training strategy achieves better anomaly classification performance.}
\item{We reformulate cross-attention module in conference paper and propose Anomaly Descriptor (AD).}
\item{We extend few-shot experiments with 8-shot setting. Meanwhile, we make more visualization to analyze normal and abnormal feature distribution, which demonstrates why samples can be classified correctly.}
\end{itemize}

\begin{figure*}
\centering
\includegraphics[width=\textwidth,]{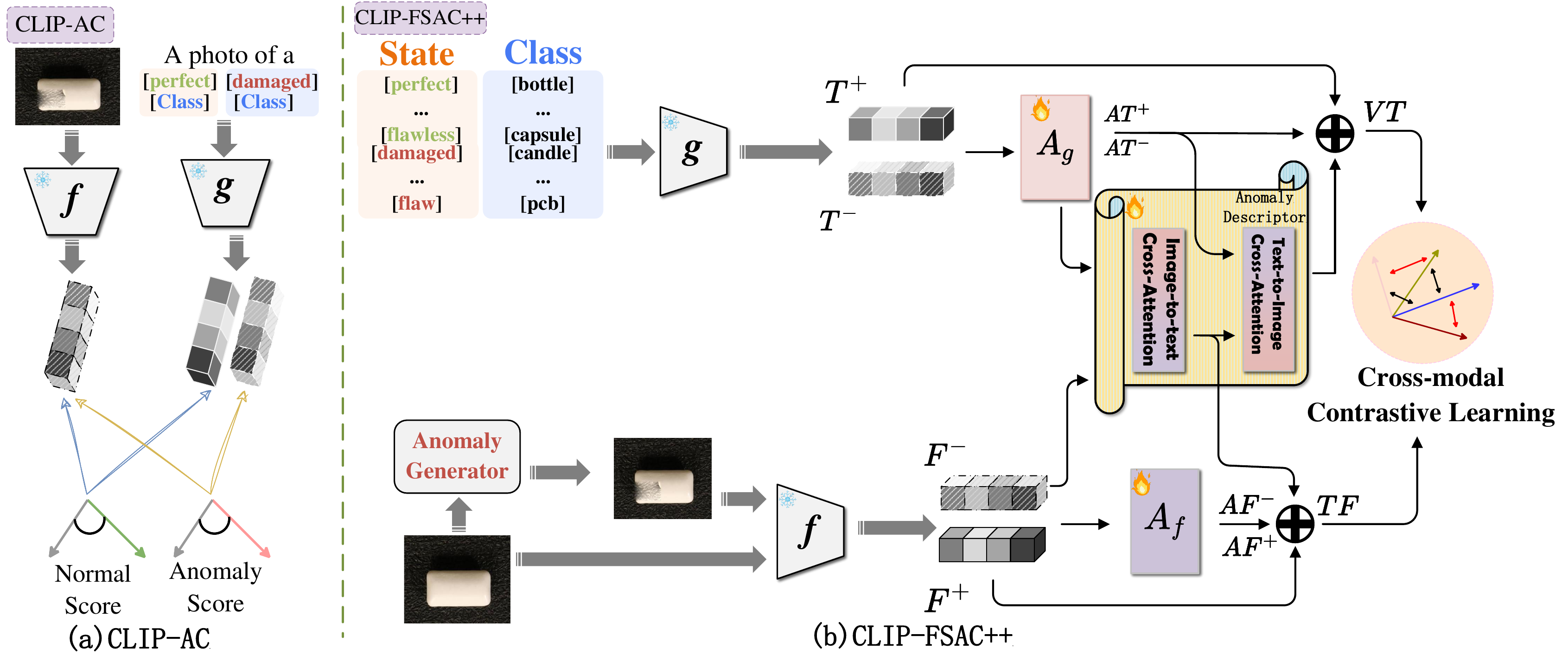}\\
\caption{The framework of CLIP-FSAC++. CLIP-AC indicates zero-shot anomaly classification with original CLIP. $f$ and $g$ are image and text encoders of CLIP, $A_f$ and $A_g$ are image and text adapters.}
\label{clipfsac++}
\end{figure*}

\section{RELATED WORK}
\subsection{Unsupervised Anomaly Detection}
For the difficulty in collecting anomalous data, most existing anomaly detection approaches rely on unsupervised learning~\cite{pni,mkd, filo, al}, which only use normal data for training models. Mainstream anomaly detection methods can be categorized into three types: reconstruction-based method~\cite{ganomaly,fd}, embedding-based method~\cite{uad} and synthesizing-based method~\cite{cutpaste}. Reconstruction-based methods employ Generative Adversarial Networks (GANs)~\cite{gan} or Diffusion Models~\cite{diad, diffusionad} to reconstruct normal image for anomaly detection. Embedding-based methods rely on a network with strong feature extraction capabilities and measure feature-level deviations to detect anomaly. Besides, synthesizing-based methods aim to synthesize anomalies on normal images so that the network can be trained with normal and abnormal samples.
Our work proposes a network which is designed to address the scarcity of samples. Our network needs few samples and has shown results that outperform the current state-of-the-art methods.
\subsection{Vision-Language Model}
Vision-language models (VLMs) have met tremendous success by exploring the interaction between textual and visual embeddings~\cite{mtvl, alphaclip, ic}. CLIP is one of the most commonly used VLMs which is trained on a dataset consisted of 400 million image-text pairs. The large amount of data and efficient designs make the pre-trained CLIP possess the strong generalization and the ability to match images and texts accurately. 
Due to the strong generality, many methods have been developed based on CLIP to guide visual tasks~\cite{pointclip, actionclip, glip}. DenseCLIP~\cite{denseclip} applies CLIP in dense prediction task. HCM~\cite{cbvrd} proposes a hierarchical context model that enriches the object-based spatial context and relation-based temporal context based on clips for video relation detection. CFine~\cite{cfine} utilize the powerful knowledge and visual representations of CLIP for ReID. FastTCM-CR50~\cite{tcm} utilizes visual prompt learning and cross-attention in CLIP to extract image and text-based prior knowledge for text detection and spotting tasks. 
WinCLIP~\cite{winclip} designs a set of prompts and proposes window-based CLIP which divides image into patches to extract dense visual features for anomaly detection. Our work uses no more than ten normal samples to fine-tune the CLIP and outperforms the state-of-the-art methods in few-shot anomaly classification.
\subsection{Few-shot Anomaly Detection}
FSAD~\cite{fastrecon, fsfa, inctrl,fewsome} is designed to adapt the practical industrial production environment. In industrial scenarios, there are few anomalous samples, and collecting images often requires human for identification. So it's difficult to get sufficient images to train the model.
In the few-shot AD setting, only few normal images are allowed to be used for training. Most FSAD researches detect anomalies by modeling the normal distribution of the few samples. Graphcore~\cite{graphcore} constructs a memory bank by features extracted from normal images based on GNN. RegAD~\cite{regad} uses meta-learning to train a category-agnostic anomaly detection model. 
Recently VLMs are popularly used in FSAD owing to their strong representative ability and generalization~\cite{clipsam, clipad}. AnomalyGPT~\cite{anomalygpt} utilize a pre-trained image encoder and a LLM to align images and their corresponding textual descriptions. PromptAD~\cite{promptad} proposes semantic concatenation and  explicit anomaly margin to solve the issues of absence of anomaly images and inaccurate prompt learning.

\section{Method}
\subsection{Preliminaries: Zero-Shot Anomaly Classification with CLIP}
Contrastive language image pre-training (CLIP) can effectively learn joint image-language representations which consist of two encoders, an image encoder $f(\cdot)$ and a text encoder $g(\cdot)$. Given an image, CLIP can correctly match images and their corresponding language representation. This ability of image-text matching can be directly used in anomaly classification. WinCLIP proposes a zero-shot anomaly classification framework named CLIP-AC illustrated in Figure 2(a). In CLIP-AC, two class prompts, $s^{+}$ =``normal [o]'' and $s^{-}$ =``anomalous [o]'' are used to identify anomaly images. [o] is a simple object-level label. For a test image, its anomaly score is defined as the cosine similarity of the image and abnormal text embeddings. To further improve performance, WinCLIP also proposes compositional prompt ensemble (CPE) including combinations of pre-defined lists of state words per label and text templates.

\subsection{Overview of CLIP-FSAC++}
Based on CLIP-FSAC, we propose CLIP-FSAC++, which further improves matching ability and generalization of CLIP in few-shot anomaly detection task. The overall framework of  CLIP-FSAC++ is illustrated in Fig. \ref{clipfsac++} (b). CLIP-FSAC++ is built upon CLIP and adapts CLIP for anomaly classification with two adapters and a cross-modality interaction module named Anomaly Descriptor (AD). There is no abnormal samples in training set. So we firstly synthesize anomalies before training using two anomalies generation methods aiming different datasets. Then normal images as positive samples and abnormal images as negative samples are sent into image encoder of CLIP to extract prior visual features. Additionally, we utilize CPE proposed in WinCLIP as text prompts. Prior text features are extracted by text encoder of CLIP. Though prior visual and text features can be matched as preliminary results owing to representative ability of CLIP demonstrated in CLIP-AC. However, directly using CLIP to classify anomaly images is subpar. We add image and text adapters following CLIP encoders to distribute representations from original CLIP to industrial domain. To enhance the correlation between vision and language, we design Anomaly Descriptor composed of two cross-modal attention modules to obtain modality-compatible features, that is, language-compatible visual features and vision-compatible text features. With less than 8 normal training samples, we transfer CLIP into FSAC task.

\begin{figure}
  \centering
  \centerline{\includegraphics[width=0.95\linewidth]{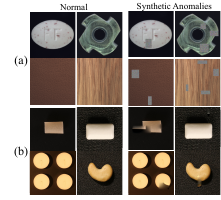}}
  \caption{Synthetic anomalies. (a) random perturbation. (b) NSA.}
  \label{noisy}
\end{figure}

\subsection{Anomalies Generation}
Due to the absence of anomalous samples in the training set, synthetic anomalies are indispensable for fine-tuning CLIP via contrastive learning. 
Initially, we generate anomalies based on given \textit{k}-shot samples like data augmentation in few-shot learning. Due to the distribution discrepancy of anomalies in different datasets, we use two anomaly synthetic methods for two different industrial datasets.
The first way to synthesize anomalies is via random perturbation~\cite{cdo}. 
Specifically, we randomly choose some square regions in normal images and replace these regions with random values sampled from a Gaussian normal distribution as shown in Fig. \ref{noisy}(a).
 The second method proposed in the NSA~\cite{nsa} appears to highlight anomalies more naturally than the first method. Natural synthetic anomalies (NSA) integrates Poisson image editing and Gamma distribution-based patch shape sampling strategy in the anomalies generation process to generate more natural and diverse anomalies as illustrated in Fig. \ref{noisy} (b).
We denote the normal image as $x^+_i$ and the synthetic anomaly image as $x^-_{i,j}$  where $(i,j)$ represents the $j$ anomaly image based on $x^+_i$. For each normal image, we will generate more than one anomaly image:
\begin{equation}
    x_{i,j}^- = Anomaly \, Generator(x_i^+).
\end{equation}

\subsection{Adapters}
For a normal training image $x^+_i$, we generate its abnormal counterpart $x^-_{i,j}$, constructing training image pairs $(x^+_i, x^-_{i,j})$. Then, these pairs are sent into CLIP image encoder $f(\cdot)$ to extract [CLS] token embeddings $F_i^+$ and $F_{i,j}^-$ as:
\begin{equation}
    F_i^+ = f(x_i^+),\ F_{i,j}^- = f(x_{i,j}^-), \quad F_i^+,F_{i,j}^-\in \mathbb{R}^{1 \times C}.
\end{equation}

In CPE, there are two sets of text prompts describing normality and anomalies respectively. We use CLIP text encoder $g(\cdot)$
to extract [EOS] tokens of each text prompt and compute average of all [EOS] tokens in normal and abnormal prompt sets:
\begin{equation}
\begin{aligned}
    T^+ = Avg(\sum_i g(s_i^+)),  T^+\in \mathbb{R}^{1 \times C}, \\
    T^- = Avg(\sum_j g(s_j^-)), T^-\in \mathbb{R}^{1 \times C},
\end{aligned}
\end{equation}
where $C$ is the token dimension. As mentioned above, the correlation between $F$ and $T$ is relatively weak. 
To improve the image-text matching ability of CLIP in anomaly classification, we introduce two adapters denoted as $A_f(\cdot)$ and $A_g(\cdot)$ for adapting visual and text representations respectively.
To avoid loss of prior information of CLIP, we blend original knowledge and adaptive knowledge via residual connection. The image adaptive feature $AF$ and text adaptive feature $AT$ are written as: 
\begin{equation}
\begin{aligned}
    AF_i^+ = \alpha_1*A_f(F_i^+)+\alpha_2*F_i^+, \, AF_i^+ \in \mathbb{R}^{1 \times C}, \\
    AF_{i,j}^- = \alpha_1*A_f(F_{i,j}^-)+\alpha_2*F_{i,j}^-, AF_{i,j}^- \in \mathbb{R}^{1 \times C},
\end{aligned}
\end{equation}

\begin{equation}
\begin{aligned}
    AT^+ = \beta_1*A_g(T^+)+\beta_2*T^+, AT^+ \in \mathbb{R}^{1 \times C}, \\
    AT^- = \beta_1*A_g(T^-)+\beta_2*T^-, AT^- \in \mathbb{R}^{1 \times C},
\end{aligned}
\end{equation}
where $\alpha$ and $\beta$ are residual ratio to balance between the original and adapted representations~\cite{clipadapter}. The relativity of visual and text embeddings is pulled closer and visual features are clustered as demonstrated in experiments.

\begin{figure}[t]
\centering
\includegraphics[width=3.33in, keepaspectratio]{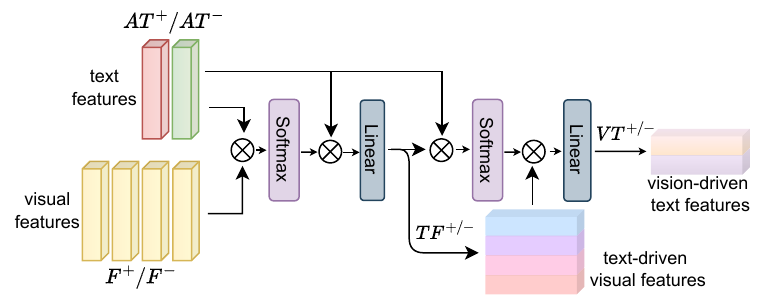}\\
\caption{Architecture of anomaly descriptor.}
\label{ad}
\end{figure}

\subsection{Anomaly Descriptor}
To further make mutually compatible between visual and text features, we design a cross-modality interaction module named Anomaly Descriptor as depicted in Fig. \ref{ad}.
Anomaly Descriptor is composed of an image-to-text cross-attention module and a text-to-image cross-attention module. Firstly, the input of image-to-text cross-attention module is original visual token embeddings $F_i^{+/-}$ and adapted text features $\psi$:
\begin{equation}
    \psi = Concat(AT^+, AT^-) ,\psi \in \mathbb{R}^{2 \times C}.
\end{equation}
Then we obtain vision-driven text features $TF^{+/-}$ by calculating the similarity of input features in a simple way:
\begin{equation}
    TF^{+/-} = Linear(Softmax(\frac{F_i^{+/-}\cdot \psi^\top}{\sqrt{C}})\psi),
\end{equation}
where $Linear(\cdot)$ is a fully-connected layer. $TF^+ \in \mathbb{R}^{1 \times C}$ is the positive vision-driven text features. $TF^- \in \mathbb{R}^{1 \times C}$ is the negative one obtained by the same way.
After obtaining vision-driven text features $TF^{+/-}$, we use $TF^{+/-}$ and adapted text features $\psi$ to calculate text-driven visual features $VT \in \mathbb{R}^{2 \times C}$ in text-to-image cross-attention module as follows:
\begin{equation}
\begin{aligned}
    TF = Concat(TF^+, TF^-), TF \in \mathbb{R}^{2 \times C} \\
    VT = Linear(Softmax(\frac{\psi \cdot TF^\mathrm{T}}{\sqrt{C}})TF). 
\end{aligned}
\end{equation}

\begin{figure}
\centering
\includegraphics[width=3.33in, keepaspectratio]{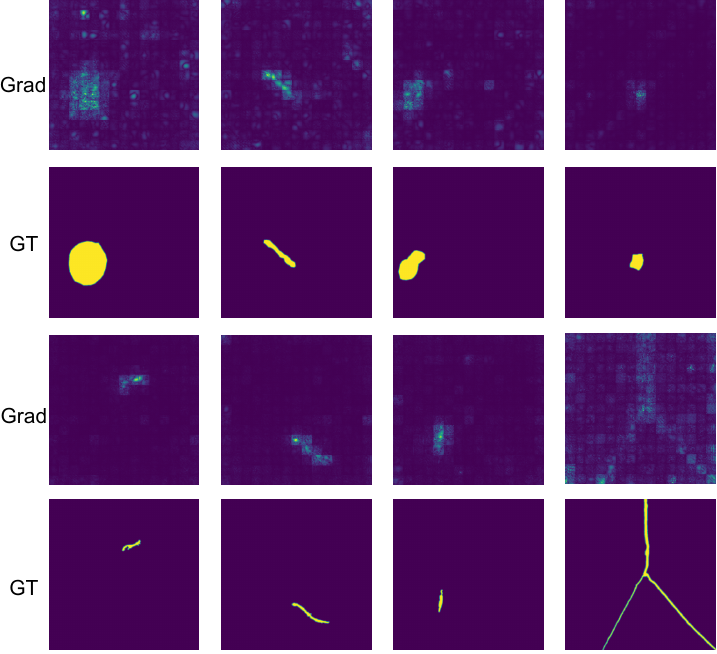}\\
\caption{Visualization of grad maps and ground truth. Yellow regions in GT denote anomalies.}
\label{grad}
\end{figure}

By attention mechanism, we obtain modality-specific driven features which mine more correlated cues between vision and language. We add text-driven visual features $TF^{+/-}$ on adapted visual embeddings $AF_i^{+/-}$ to generate text-enhanced visual features $CF_i^{+/-}$ and add vision-driven text features $VT$ on adapted text embeddings $\psi$ to generate vision-enhanced text features denoted as $CT$:
\begin{equation}
\begin{aligned}
    CT = \psi + \gamma_1 * VT, \, CT \in \mathbb{R}^{2 \times C} \\ 
    CF_i^+ = AF_i^+ + TF^+, CF_i^+ \in \mathbb{R}^{1 \times C}\\ 
    CF_i^- = AF_i^- + TF^-,   CF_i^- \in \mathbb{R}^{1 \times C},
\end{aligned}
\end{equation}
where $\gamma_1$ is a hyper-parameter which controls the proportion of 
vision-driven text features in final text features.

Anomaly Descriptor dramatically facilitates the vision-language matching ability of CLIP in FSAC and improves few-shot anomaly classification performance as demonstrated in Tab. \ref{aad}.

\subsection{Loss}
We denote the training set and testing set as $\chi_{train}$ including less than 8 normal samples and $\chi_{test}$. Instead of two-stage strategy in CLIP-FSAC, we optimize all modules jointly in one stage. The loss for optimization consists of three parts: image-to-text contrastive loss, text-to-image contrastive loss and cross-entropy loss~\cite{moco}. Image-to-text contrastive loss $\mathcal{L}_{i2t}$ is calculated as: 
\begin{equation}
\begin{aligned}
\mathcal{L}_{i2t} = -log\frac{exp(s(CF_i^+, AT^+))}{exp(s(CF_i^+, AT^+))+exp(s(CF_i^+, AT^-))}\\
-log\frac{exp(s(CF_{i,j}^-, AT^-))}{exp(s(CF_{i,j}^-, AT^-))+exp(s(CF_{i,j}^-, AT^+))},
\end{aligned}
\end{equation}
where $s(\cdot,\cdot)$ is consine similarity function. The second part is text-to-image contrastive loss $\mathcal{L}_{t2i}$. It is calculated as:
\begin{equation}
\begin{aligned}
\mathcal{L}_{t2i} = -log\frac{exp(s(CF_i^+, AT^+))}{exp(s(CF_i^+, AT^+))+exp(s(CF_{i,j}^-, AT^+))}\\
-log\frac{exp(s(CF_{i,j}^-, AT^-))}{exp(s(CF_{i,j}^-, AT^-))+exp(s(CF_i^+, AT^-))}.
\end{aligned}
\end{equation}
The final part is cross-entropy loss which is calculated as:
\begin{equation}
\begin{aligned}
\mathcal{L}_{ce} = CrossEntropy(s(CF, CT^-), Label).
\end{aligned}
\end{equation}
where $s(CF, CT^-)$ is consine similarity between visual features and abnormal text features. $Label$ is the ground truth label of samples:
\begin{equation*}
{Label} =
\begin{cases}
1,&{\text{if}}\ sample \,is \,abnormal,\\
{0,}&{\text{if}}\ sample \,is \,normal,
\end{cases}
\end{equation*}
In a word, the total loss for training is
\begin{equation}
\mathcal{L}_{total} = \mathcal{L}_{t2i} + \mathcal{L}_{i2t} + \gamma_2 * \mathcal{L}_{ce},
\end{equation}
where $\gamma_2$ is loss hyperparameter and controls the proportion of cross-entropy loss in total loss.

\begin{table*}[h]
\caption{Anomaly Classification Performance Evaluated by The Metric of I-AUROC on The VisA Dataset. Bold Presents Optimal Results.}
\label{iaurocvisa}
\centering
\resizebox{0.98\linewidth}{!}{
\begin{tabular}{cl|cccccccccccc|c}
\hline
\multicolumn{2}{c|}{Method}&candle &cashew & capsules & chewinggum & fryum & macaroni1 & macaroni2 & pcb1 & pcb2 & pcb3 & pcb4 & pipe\_fryum& mean \\ \hline
& Padim \cite{padim} & 70.8 & 62.3 & 51.0 & 69.9 & 58.3 & 62.1 & 47.5 & 76.2 & 61.2 & 51.4 & 76.1 & 66.7 & 62.8 \\
& Patchcore \cite{patchcore} & 85.1  & 89.5 & 60.0 & 97.3 & 75.0 & 68.0 & 55.6 & 78.9 & 81.5 & 82.7 & 93.9 & 90.7 & 79.9\\
1-shot
& SPADE \cite{spade} & 86.1 & 95.9 & 73.3 & 92.1 & 81.1  & 66.0 & 55.8 & 87.2 & 73.5 & 72.2 & 93.4 & 77.9 & 79.5\\
& WinCLIP~\cite{winclip}  & 93.4  & 94.0 & 85.0 & 97.6 & 88.5 & 82.9 & 70.2 & 75.6 & 62.2 & 74.1 & 85.2 & 97.2 & 83.8\\
& CLIP-FSAC~\cite{clipfsac}  & \textbf{100}  & 99.9 & 78.6 & 96.7 & 99.1 & 92.3 & 87.9 & 99.8 & 89.9 & 97.3 & 99.3 & 90.7 & 96.0\\
& CLIP-FSAC++(Ours) & 99.9 & \textbf{99.7} & \textbf{94.6} & \textbf{100} & \textbf{96.2}  & \textbf{97.5} & \textbf{97.8} & \textbf{99.5} & \textbf{93.6} & \textbf{98.3} & \textbf{99.0} & \textbf{96.1} & \textbf{97.6} \\ \hline
& Padim \cite{padim} & 75.8  & 74.6 & 51.7 & 82.7 & 69.2 & 62.2 & 50.8 & 62.4 & 66.8 & 67.3 & 69.3 & 75.3 & 67.4 \\
& Patchcore \cite{patchcore} & 85.3  & 93.6 & 57.8 & 97.8 & 83.4 & 75.6 & 57.3 & 71.5 & 84.3 & 84.8 & 94.3 & 93.5 & 81.6 \\
2-shot
& SPADE \cite{spade} & 91.3 & 97.3 & 71.7 & 93.4 & 90.5 & 69.1 & 58.3 & 86.7  & 70.3 & 75.8 & 86.1  & 78.1 & 80.7\\
& WinCLIP \cite{winclip}& 94.8 & 94.3 & 84.9 & 97.3 & 90.5 & 83.3 & 71.8 & 76.7 & 62.6 & 78.8 & 82.3 & 98.0 & 84.6\\
& CLIP-FSAC~\cite{clipfsac}  & \textbf{100}  & 99.9 & 93.2 & 100 & 98.9 & 98.0 & 82.3 & 99.9 & 80.3 & 98.0 & 99.7 & 96.1 & 95.5\\
&CLIP-FSAC++(Ours) & \textbf{99.8} & \textbf{99.5} & \textbf{93.8} & \textbf{100} & \textbf{96.2} & \textbf{99.1} & \textbf{96.1} & \textbf{100} & \textbf{91.6}  & \textbf{98.1} &  \textbf{99.4} & \textbf{95.8} &  \textbf{97.4}  \\ \hline
& Padim \cite{padim}& 77.5 & 77.7 & 52.7 & 83.5 & 71.2 & 65.9 & 55.0 & 82.6  & 73.5 & 65.9 & 85.4 & 82.9 & 72.8\\
& Patchcore \cite{patchcore}& 87.8 & 93.0 & 63.4 & 98.3 & 88.6 & 82.9 & 61.7 & 84.7  & 84.3 & 87.0 & 95.6 & 96.4 & 85.3 \\
4-shot
& SPADE \cite{spade}  & 92.8 & 96.4 & 73.4 & 93.5 & 92.9 & 65.8 & 56.7 & 83.4 & 71.7 & 79.0 & 95.4 & 79.3 & 81.7\\
& WinCLIP \cite{winclip}& 95.1 & 95.2 & 86.8 & 97.7 & 90.8 & 85.2 & 70.9  & 88.3 & 67.5 & 83.3 & 87.6  & 98.5 &87.3\\
& CLIP-FSAC~\cite{clipfsac}  & 99.9  & 99.7 & 94.4 & 100 & 98.7 & 100 & 95.4 & 99.9 & 93.1 & 95.6 & 99.7 & 96.2 & \textbf{97.7}\\
& CLIP-FSAC++(Ours) & \textbf{99.9} & \textbf{99.7} & \textbf{94.5} & \textbf{100} & \textbf{97.5} & \textbf{99.4} & \textbf{97.0} & \textbf{99.2} & \textbf{92.5} & \textbf{96.7}  & \textbf{99.4}  & \textbf{95.0} & 97.6  \\ 
\hline
& Padim \cite{padim}& 68.8 & 77.2 & 58.2 & 88.6 & 77.8 & 62.3 & 59.5 & 73.8  & 72.1 & 49.4 & 84.6 & 83.7 & 71.3\\
\multirow{2}*{8-shot}
& Patchcore \cite{patchcore}& 93.8 & 95.30 & 63.7 & 98.6 & 89.7 & 78.7 & 70.4 & 77.2  & 88.6 & 86.9 &  98.3&  97.8& 86.6 \\

& WinCLIP \cite{winclip}& 96.9 & 95.6 & 82.6 & 97.9 & 89.7 & 89.6 & 76.7  & 87.4 & 63.7 & 76.1 &84.6 &91.2& 86.0\\
& CLIP-FSAC++(Ours) & \textbf{99.9} & \textbf{99.4} & \textbf{94.6} & \textbf{100} & \textbf{98.1} & \textbf{98.0} &  \textbf{94.0} & \textbf{99.9}  & \textbf{95.9} & \textbf{98.5}  & \textbf{98.2} & \textbf{95.8} & \textbf{97.7}   \\ 
\hline
\end{tabular}%
}
\end{table*}

\begin{table*}[h]

\caption{Anomaly Classification Performance Evaluated by The Metric of I-AUROC on The MVTEC-AD Dataset. Bold Presents Optimal Results.}
\label{iaurocmvtec}
\centering
\resizebox{\linewidth}{!}{
\begin{tabular}{cl|ccccccccccccccc|c}
\hline
\multicolumn{2}{c|}{Method}& carpet & bottle  & hazelnut & leather & cable & capsule & grid & pill & transistor & metal\_nut& screw &  toothbrush &  zipper & tile & wood & mean \\ \hline
& Padim \cite{padim} & 96.6 & 97.4 & 88.3 & 97.5 & 57.7 & 57.7 & 54.2 & 61.3 & 73.3 & 53.0 & 55.0 & 82.5 & 85.8 & 92.2 & 96.1 & 76.6\\
& Patchcore \cite{patchcore} & 95.3  & 99.4 & 88.3 & 97.3 & 88.8 & 67.8 & 63.6 & 81.9 & 78.1 & 73.4 & 44.4 & 83.3 & 92.3 & 99.0 & 97.8 & 83.4\\
1-shot
& GraphCore \cite{graphcore} & 99.3 & 99.8 & 98.5 &  \textbf{100} & 91.1  & 72.1 & 80.9 & 81.2 & 96.2 & 92.5 & 57.9 & 85.2 & 97.5 & 99.2 & 97.3 & 89.9\\
& WinCLIP \cite{winclip} & 99.8  & 98.2 & 97.5 & 99.9 & 88.9 & 72.3 & 99.5 & 91.2 & 83.4 & 98.7 & 86.4 & 92.2 & 88.8 & 99.9 & 99.9& 93.1\\
& CLIP-FSAC \cite{clipfsac} & 100  & 96.4 & 97.1 & 100 & 84.4 & 93.7 & 99.8 & 83.1 & 91.5 & 93.5 & 95.2 & 99.4 & 98.2 & 100 & 100 & \textbf{95.5}\\
& CLIP-FSAC++(Ours) & \textbf{100} & 98.4 & \textbf{98.8} & \textbf{100} & \textbf{96.7} & \textbf{98.6}  & \textbf{99.5} & 70.8 & \textbf{99.8} & \textbf{98.2} & 55.1 & \textbf{99.4} & \textbf{97.7} & \textbf{100} & 97.2& 94.0  \\ \hline
& Padim \cite{padim} & 97.8  & 98.5 & 90.8 & 97.5 &62.3& 64.3 & 67.2 & 59.1 & 72.8 & 54.8 & 54.0 & 87.6 & 86.3 & 93.3 & 96.9 & 78.9 \\
& Patchcore \cite{patchcore} & 96.6  & 99.2 & 93.2 & 97.9 & 91.0 & 72.8 & 67.7 & 82.9 & 90.0 & 77.7 & 49.0 & 85.9 & 94.0 & 98.5 & 98.3 & 86.3\\
2-shot
& GraphCore \cite{graphcore} & 99.4 & 99.8 & \textbf{99.5} & 100 & 95.2 & 73.2 &  81.5 & 88.6  & 97.1 & 96.3 & 65.7 & 87.3 & 97.5  & \textbf{100} & 97.5 & 91.91\\
& WinCLIP \cite{winclip} & 99.8 &99.3 & 98.3 & 99.9 & 88.4 & 77.3 & 99.4 & 92.3 & 85.3 & 99.4 & 86.0 & 97.5 & 94.0  & 99.9 & 99.9 & 94.4\\
& CLIP-FSAC \cite{clipfsac} & 100  & 99.3 & 96.5 & 99.8 & 84.4 & 91.6 & 100 & 77.8 & 61.4 & 96.0 & 84.8&99.4 & 97.7 & 100 & 100 & 92.6 \\

& CLIP-FSAC++(Ours) & \textbf{100} & 98.8 & \textbf{99.5} & \textbf{100} & \textbf{96.6} &  \textbf{97.6} & \textbf{100} & 82.2 & 95.5  & 98.7 & 81.1 & \textbf{100} & \textbf{99.2}& \textbf{100} & 96.1 & \textbf{96.3} \\ \hline
& Padim \cite{padim}& 97.9 & 98.8 & 91.9 & 98.5 & 70.0 & 65.2 & 68.1 & 54.9  &82.4 & 60.7 & 50.0 & 89.2  & 88.3 & 89.2 & 97.0 & 80.4 \\
& Patchcore \cite{patchcore}& 96.6 & 99.2 & 93.2 & 97.9 & 91.0 & 72.8 & 67.7 & 82.9  & 90.0 & 77.7 & 49.0 & 85.9  & 94.0 & 98.5 & 98.3& 88.8\\
4-shot
& GraphCore \cite{graphcore} & 99.4 & 99.8 & 99.5 & 100 & 95.2 & 74.5 & 81.6 & 88.2 & 99.2 & 96.2 & 68.9 & 95.2 & 98.2 & 100 & 97.9& 92.9 \\
& WinCLIP \cite{winclip} & 100.0 & 99.3 & 98.4 & 100.0 & 90.9 & 82.3 & 99.6  & 92.8 & 85.7 & 99.5 & 87.9  & 96.7 &94.5 & 99.9 &99.8 & 95.2\\
& CLIP-FSAC \cite{clipfsac} & 100.0 & 100.0 & 97.4 & 100.0 & 84.5 & 91.9 & 99.8  & 76.6 & 93.6 & 97.5 &98.1  & 99.4 &98.4 & 100& 99.8 & 95.8\\
& CLIP-FSAC++(Ours) & \textbf{100} & \textbf{99.8} & 99.0 & 99.9 & \textbf{96.6} & \textbf{97.6} & \textbf{99.9} & 75.8  & 98.8& 98.8& 81.0  & \textbf{99.2} & \textbf{99.3}& \textbf{100} & 98.2 & \textbf{96.3} \\ 
\hline
& Padim \cite{padim}& 99.8 & 98.3 & 95.5 & 99.9 & 68.0 & 77.2 & 67.8 & 59.2  & 69.8 & 81.0 & 54.8 & 87.2  & 85.2 & 94.6 & 98.1 & 82.4\\
& Patchcore \cite{patchcore}& 98.9 & 100 & 98.7 & 100 & 89.7 & 91.3 & 85.5 & 91.0  & 87.4 & 97.0 & 61.9 & 82.5 & 97.8 & 99.1 & 98.5 & 92.0\\
8-shot
& GraphCore \cite{graphcore} & 98.5 & 99.8 & \textbf{100} & \textbf{100} & 95.2 & 90.5 & 92.3 & 91.1 & 99.2 & 97.9 &  80.1 & 95.1 & 99.2 & 100 & 98.9 & \textbf{95.9}\\
& WinCLIP \cite{winclip} & 99.8 & 99.5 & 98.4 & \textbf{100} & 91.4 & 85.5 & 99.6  & 93.3 & 91.1 & 99.2 &80.4 &98.9	&97.6&\textbf{100}	&99.6& 95.6\\
& CLIP-FSAC++(Ours) & \textbf{100} & 98.4 & 99.3 & \textbf{100} & 95.8 & \textbf{98.6} & \textbf{100} &69.2  & 95.8& 98.3  & 70.4 & \textbf{99.4} & 94.6&  \textbf{100} & 98.1 & 94.5 \\ 
\hline
\end{tabular}%
}
\end{table*}

\subsection{Anomaly Classification with CLIP-FSAC++}
In testing phase, synthetic anomalies are no longer needed. Testing image $x_{test} \in \chi_{test}$ is fed into CLIP image encoder to extract visual features, which is adapted and enhanced. Two vision-enhanced text features $CT^+$ and $CT^-$ representing normality and abnormity are generated by CLIP text encoder and proposed modules. Then we calculate positive score $S^+$ and negative score $S^-$, also known as normal score and anomaly score, for given testing image $x_{test}$ as follows:
\begin{equation}
S^+(x_{test}) = \frac{exp(\frac{s(CF_{test}, CT^+)}{\tau})}{exp(\frac{s(CF_{test}, CT^-)}{\tau})+exp(\frac{s(CF_{test}, CT^+)}{\tau})},
\end{equation}
\begin{equation}
S^-(x_{test}) = \frac{exp(\frac{s(CF_{test}, CT^-)}{\tau})}{exp(\frac{s(CF_{test}, CT^-)}{\tau})+exp(\frac{s(CF_{test}, CT^+)}{\tau})},
\end{equation}
where $\tau$ is a temperature hyper-parameter.
Finally, we calculate the final anomaly score as
\begin{equation}
AS(x_{test}) = \frac{S^-}{S^-+S^+}.
\end{equation}

As mentioned in CLIP-FSAC, we calculate the gradient map as follows:
\begin{equation}
grad = \left| \frac{ \partial AS(x_{test}) }{ \partial x_{test}}\right| \in \mathbb{R}^{H \times W \times 3}
\end{equation}
Then we average $grad$ along channel dimension to obtain $grad\_map$, which represents the contribution of each region in input image to final anomaly score $AS(x_{test})$. As illustrated in Fig. \ref{grad}, it is demonstrated that anomaly regions contribute more to final anomaly score $AS(x_{test})$ than normal regions. This elucidates why testing samples can be classified correctly and the matching and representative ability of CLIP is improved a lot to achieve a good performance in few-shot anomaly classification.

\begin{figure}
\centering
\includegraphics[width=3.0in, keepaspectratio]{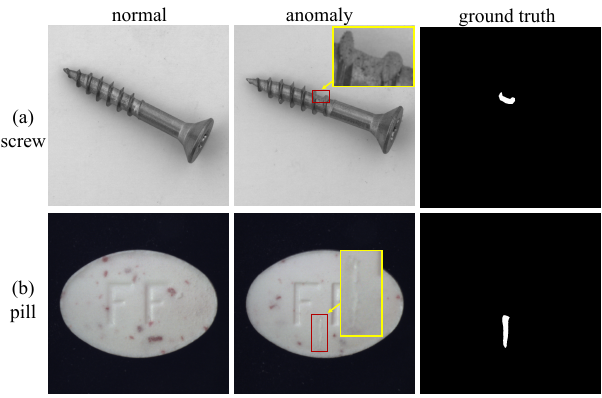}\\
\caption{Examples of hard samples and anomalies.}
\label{hard}
\end{figure}

\section{EXPERIMENTS}
\subsection{Experimental Setting}
\subsubsection{Datasets}
To evaluate the effectiveness of our proposed method,
we conducted extensive experiments on two commonly used datasets, MVTEC-AD~\cite{mvtec} and VisA~\cite{visa}. Notably, there are only normal images in training set of these two datasets. The details of the datasets are as follows: 
\begin{itemize}
\item{\textbf{MVTEC-AD} is the most commonly used anomaly detection dataset with  totally 5354 high-resolution images. It consists of 15 classes including 5 texture categories and 10 object categories such as bottle, pill, grid and so on. There are totally 73 kinds of anomalies in MVTEC-AD. The number of training samples for each category falls in the area between 60 to 320.}
\item{\textbf{VisA} is also a commonly used anomaly detection dataset which is larger than MVTec-AD. It collects 10,821 high-resolution RGB images, consisting of 12 different classes across 3 domains including candle, pcb, fryum, etc. }
\end{itemize}

\begin{table*}[ht]
\setlength{\tabcolsep}{5pt}
  \centering
  \caption{Anomaly Classification Performance Comparison of The Proposed CLIP-FSAC++ Against The SOTA Approaches on The VisA And MVTEC-AD Benchmarks.  Red Indicates The Best Result, And Blue Displays The Second-best.}
  \label{all}
  \resizebox{0.98\linewidth}{!}{
    \begin{tabular}{cc|c|llllll}
    \hline
    \multirow{2}*{Setup} & \multicolumn{1}{c}{\multirow{2}*{Method }} & \multicolumn{1}{c}{\multirow{2}*{Venue}} & \multicolumn{3}{c}{VisA} & \multicolumn{3}{c}{MVTEC-AD} \\
\cmidrule{4-9}          & \multicolumn{1}{c}{} & \multicolumn{1}{c}{} & I-AUROC & I-AUPR & F1-MAX & I-AUROC & I-AUPR & F1-MAX \\
    \hline
          & PaDiM \cite{padim} & ICPRW2021 & 62.8  & 68.3  & 75.3  & 76.6  & 88.1  & 88.2  \\
          & PatchCore \cite{patchcore} & CVPR2022 & 79.9  & 82.8  & 81.7  & 83.4  & 92.2  & 90.5  \\
          & RegAD \cite{regad} & ECCV2022 & -  & -  & -  & 82.4  & -  & -  \\
    \multirow{2}*{1-shot}      & GraphCore \cite{graphcore} & ICLR2023 &   -    &    -   &   -    & 89.9  &   -    & - \\
          & WinCLIP \cite{winclip} & CVPR2023 & 83.8  & 85.1  & 83.1  & 93.1  & \textcolor{blue}{96.5}  & \textcolor{blue}{93.7}  \\
          & APRIL-GAN \cite{aprilgan} & CVPRW2023 & \textcolor{blue}{91.2}  & \textcolor{blue}{93.3}  & \textcolor{blue}{86.9}  & 92.0  & 95.8  & 92.4  \\
          & AnomalyGPT \cite{anomalygpt} & AAAI2024 & 87.4  & -  & - & 94.1  & -  & -  \\
          & AnoPLe \cite{anople} & 2024 & 86.2  & -  & -  & 94.1  & -  & -  \\
          & CLIP-FSAC \cite{clipfsac} & IJCAI2024 & 96.0  & 96.0  & 93.4  & \textcolor{red}{95.5}  & 97.5  & 95.0  \\
\cmidrule{2-9}          
          & CLIP-FSAC++ (ours) &      & \textcolor{red}{97.6}  & \textcolor{red}{98.3}  & \textcolor{red}{94.7}  & \textcolor{blue}{94.0}  & \textcolor{red}{98.1}  & \textcolor{red}{96.2}  \\
    \hline
          & PaDiM \cite{padim} & ICPRW2021 & 67.4  & 71.6  & 75.7  & 78.9  & 89.3  & 89.2  \\
          & PatchCore \cite{patchcore} & CVPR2022 & 81.6  & 84.8  & 82.5  & 86.3  & 93.8  & 92.0  \\
          & RegAD \cite{regad} & ECCV2022 & -  & -  & -  & 85.7  & -  & -  \\
    \multirow{2}*{2-shot} & GraphCore \cite{graphcore} & ICLR2023 &    -   &    -   &    -   & 91.9  &    -   & - \\
          & WinCLIP \cite{winclip} & CVPR2023 & 84.6  & 85.8 & 83.0 & 94.4  & \textcolor{blue}{97.0}  & \textcolor{blue}{94.4}  \\
          & APRIL-GAN \cite{aprilgan} & CVPRW2023 & 92.2  & 94.2  & 87.7   & 92.4  & 96.0  & 92.6  \\
          & AnomalyGPT \cite{anomalygpt} & AAAI2024 & 88.6  & -  & - & 95.5  & -  & -  \\
          & AnoPLe \cite{anople} & 2024 & 86.6  & -  & -  & \textcolor{blue}{95.3}  & -  & -  \\
          & CLIP-FSAC \cite{clipfsac} & IJCAI2024 & \textcolor{blue}{95.5}  & \textcolor{blue}{95.5}  & \textcolor{blue}{93.3}  & 92.6  & 95.4  & 93.0  \\
\cmidrule{2-9}         
          & CLIP-FSAC++ (ours) &       & \textcolor{red}{97.4}  & \textcolor{red}{98.2}  & \textcolor{red}{94.6}  & \textcolor{red}{96.3}  & \textcolor{red}{98.6}  & \textcolor{red}{96.4}  \\
    \hline
          & PaDiM \cite{padim} & ICPRW2021 & 72.8  & 75.6  & 78.0  & 80.4  & 90.5  & 90.2  \\
          & PatchCore \cite{patchcore} & CVPR2022 & 85.3  & 87.5  & 84.3  & 88.8  & 94.5  & 92.6  \\
         & RegAD \cite{regad} & ECCV2022 & -  & -  & -  & 88.2  & -  & -  \\
    
    \multirow{2}*{4-shot}  & GraphCore \cite{graphcore} & ICLR2023 &   -    &    -   &    -   & 92.9  &   -    & - \\
          & WinCLIP \cite{winclip} & CVPR2023 & 87.3  & 88.8  & 84.2 & 95.2  & 97.3  & 94.7  \\
          & APRIL-GAN \cite{aprilgan} & CVPRW2023 & 92.6  & 94.5  & 88.4  & 92.8  & 96.3  & 92.8  \\
          & AnomalyGPT \cite{anomalygpt} & AAAI2024 & 90.6  & -  & - & \textcolor{blue}{96.3}  & -  & -  \\
          & AnoPLe \cite{anople} & 2024 & 87.7  & -  & -  & \textcolor{blue}{96.3}  & -  & -  \\
          & CLIP-FSAC \cite{clipfsac} & IJCAI2024 & \textcolor{red}{97.7}  & \textcolor{blue}{97.9}  & \textcolor{red}{95.5}  & 95.8  & \textcolor{blue}{98.0}  & \textcolor{blue}{95.6}  \\
\cmidrule{2-9}          
          & CLIP-FSAC++ (ours) &       & \textcolor{blue}{97.6}  & \textcolor{red}{98.2}  & \textcolor{blue}{94.6}  & \textcolor{red}{96.3}  & \textcolor{red}{98.7}  & \textcolor{red}{96.7}  \\
    \hline
     & PaDiM \cite{padim} & ICPRW2021 & 71.3 & 73.2  & 78.1  & 82.4  & 90.7  & 90.4  \\
          & PatchCore \cite{patchcore} & CVPR2022 & 86.6  & 88.1  & 85.6  & 92.0  & 95.9  & 93.3  \\
         & RegAD \cite{regad} & ECCV2022 & -  & -  & -  & 91.2  & -  & -  \\
    
    \multirow{2}*{8-shot}  & GraphCore \cite{graphcore} & ICLR2023 &   -    &    -   &    -   & 95.9  &   -    & - \\
          & WinCLIP \cite{winclip} & CVPR2023 & 86.0  & 87.1  & 83.8  & 95.6  & 97.6 & 95.4  \\
\cmidrule{2-9}          
          & CLIP-FSAC++ (ours) &       & \textcolor{red}{97.7}  & \textcolor{red}{98.3}  & \textcolor{red}{94.5}  & \textcolor{red}{94.5}  & \textcolor{red}{98.0}  & \textcolor{red}{95.4}  \\
    \hline
\end{tabular}
}%
\end{table*}%

\subsubsection{Evaluation Metrics}
Our few-shot anomaly classification performance is evaluated based on three metrics following previous anomaly detection methods: (1) Area Under the Receiver Operating Characteristic (I-AUROC) which is the most commonly used metric for anomaly detection tasks, (2) Area Under the Precision-Recall
curve (I-AUPR) which is proposed to address imbalance issues and (3) F1-Max which measures F1-score for anomaly classification at optimal threshold.

\subsubsection{Implementation Details}
CLIP image and text encoders are realized by OpenCLIP~\cite{openclip} and directly load pre-trained CLIP checkpoints called LAION-400M~\cite{laion400}. In anomalies generation phase, we use random perturbation to generate anomalies on Mvtec AD and NSA on VisA. Our image adapter consists of two multi-layer perceptrons (MLPs) and text adapter is composed of one MLP.
We use Adam optimizer with learning rate of 0.0005 for image adapter and 0.0001 for text adapter. Training epoch is set to 100 for all datasets. To conduct thorough few-shot experiments, our few-shot anomaly classification settings are set to 1-shot, 2-shot, 4-shot and 8-shot. Batch size is set to 1,1,2,2 and 1,2,2,2 for VisA and MVTEC-AD respectively. Hyperparameters $\alpha_1$ and $\alpha_2$ in Equation 4 are set to 0.4, 0.6 and 0.1, 0.9 for MVTEC-AD and VisA.  Hyperparameters $\beta_1$ and $\beta_2$ in Equation 5 are also set 0.4, 0.6 and 0.1, 0.9 for MVTEC-AD and VisA datasets. Besides, $\gamma_1$ is set to 0.7 for VisA and 0.01 for MVTEC-AD. $\gamma_2$ is set to 0.7 for both datasets. Adam optimizer is used for training. CLIP-FSAC++ is deployment under PyTorch on single NVIDIA RTX 3090.

\begin{table}[t]
\caption{Quantitative Comparison (I-AUROC) Between CLIP-FSAC++ And Full-shot Methods on VisA And MVTEC-AD Datasets.}
 \label{full}
    \centering
    \begin{tabular}{cccc}
       \hline
        Method  & k-shot & VisA & MVTEC \\
        \hline
        CLIP-FSAC++ (ours)     & 1-shot     & 97.6  &94.0 \\
        CLIP-FSAC++ (ours)     & 2-shot     & 97.4  & 96.3  \\
        CLIP-FSAC++ (ours)  & 4-shot     & 97.6  &96.3 \\
        CLIP-FSAC++ (ours)  & 8-shot     & 97.7  &94.5 \\
        \hline
        \thead{EfficientAD-S \cite{efficientad}} & full-shot   & 97.5 & 98.8 \\
        FAIR \cite{fair} & full-shot    & 96.7 & 98.6\\
        EdgRec \cite{edgrec} & full-shot    & 94.2 & 97.8 \\  
        CutPaste \cite{cutpaste} & full-shot  &  - & 95.2 \\ 
       MKD \cite{mkd} & full-shot     & - &87.7 \\ 
        DiffusionAD \cite{diffusionad} & full-shot     & 98.8 &99.7 \\
        GLASS \cite{glass} & full-shot     & 98.8 &99.9 \\
        \hline
    \end{tabular}
    \label{tab:plain}
\end{table}

\begin{table}[ht]
  \centering
  \caption{Ablation Studies on Different Methods of Loss Function. (I-AUROC/I-AUPR/F1-MAX)}
 \label{lf}
    \begin{tabular}{cccccc}
    \hline
       $\mathcal{L}_{t2i} + \mathcal{L}_{i2t}$ & $\mathcal{L}_{ce}$  & Setup & VisA  & MVTEC-AD \\
    \hline
    &  & K=1   & 96.3/96.6/93.7 & 93.7/98.0/95.1 \\
    \multirow{2}*{\usym{2714}}  & \multirow{2}*{\usym{2717}} & K=2   & 97.6/98.3/95.0 & 96.1/98.6/95.5 \\
        &  & K=4   & 96.4/96.9/93.6 & 93.3/98.0/96.1 \\
        &   & K=8   & 96.8/97.4/94.3 & 93.6/98.1/95.8 \\
    \hline
    & & K=1   & 96.7/97.3/93.8 & 94.9/98.3/95.7 \\
   \multirow{2}*{\usym{2717}}  & \multirow{2}*{\usym{2714}} & K=2   & 95.8/96.5/93.0 & 94.8/98.4/96.3 \\
      &    & K=4   & 96.8/97.3/94.0 & 96.2/98.6/96.3 \\
       &    & K=8   & 97.5/98.2/94.3 & 94.5/98.3/96.4 \\
    \hline
    & & K=1   & 97.6/98.3/94.7 & 94.0/98.1/96.2 \\
    \multirow{2}*{\usym{2714}} & \multirow{2}*{\usym{2714}} & K=2   & 97.4/98.2/94.6 & 96.3/98.6/96.4 \\
      &    & K=4   & 97.6/98.2/94.6 & 96.3/98.7/96.7 \\
       &    & K=8   & 97.7/98.3/94.5 & 94.5/98.0/95.4 \\
    \hline
    \end{tabular}%
    
  \label{tab:addlabel}%
\end{table}%

\subsection{Quantitative Analysis}
\subsubsection{Comparison with the State-of-the-art}
In this section, we report our detailed results of each class on MVTEC-AD and VisA datasets shown in Tab.~\ref{iaurocvisa} and Tab.~\ref{iaurocmvtec} about I-AUROC. We compare our anomaly classification performance with previous state-of-the-art anomaly detection methods in four few-shot setups. In the few-normal-shot setup, our CLIP-FSAC++ achieves new SOTA of 97.6\%, 97.4\%, 97.6\% and 97.7\%of I-AUROC on VisA for 1-shot, 2-shot, 4-shot and 8-shot respectively, improving upon the state-of-the-art WinCLIP\cite{winclip} by 13.8\%, 12.8\%,d 10.3\% and 11.7\% considerably. As shown in Tab.~\ref{iaurocvisa}, CLIP-FSAC++ achieves the best performance on almost all of classes in VisA dataset. For MVTEC-AD dataset, CLIP-FSAC++ reaches competitive 94.0\%, 96.3\% 96.3\% and 94.5\% of I-AUROC on MVTEC-AD for 1-shot, 2-shot, 4-shot and 8-shot showed in Tab.~\ref{iaurocmvtec}. We note that CLIP-FSAC++ reaches the highest I-AUROC in 2-shot and 4-shot while the performance drops in 8-shot. We analyze the reasons why performance drops in 8-shot and draw a conclusion: the quality of synthetic anomalies when training in 8-shot is worse, which misleads the model to unexpected vision-language matching. Meanwhile, we find that the results of "screw" and "pill" as hard classes in MVTEC-AD are not satisfactory and stable. We analyze this is because anomalies in these two classes are small and indistinguishable as depicted in Fig.~\ref{hard}. It can be seen that anomalies in these two classes are not distinct so that it can not be classified as anomaly sample correctly. To measure the performance of our method comprehensively, we also use two other metrics I-AUPR and F1-Max to demonstrate the classification results except I-AUROC. As shown in Tab.~\ref{all}, we compare our method CLIP-FSAC++ with many previous SOTA methods in I-AUPR and F1-Max on MVTEC-AD and VisA datasets.
Because there exists randomness in anomalies generation, our experiments run five times using different random seeds and we also report the corresponding standard error.

\begin{table}[t]
  \centering
  \caption{Ablation Studies on Different Methods of Anomaly Descriptor. (I-AUROC/I-AUPR/F1-MAX)}
  \label{aad}
    \begin{tabular}{cccccc}
    \hline
       $TF$ & $VT$  & Setup & VisA  & MVTEC-AD \\
       \hline
    &  & K=1   & 72.5/75.8/77.8 & 88.3/95.8/92.7 \\
    \multirow{2}*{\usym{2717}}  & \multirow{2}*{\usym{2717}} & K=2   & 64.4/69.4/76.1 & 90.3/96.0/92.9 \\
        &  & K=4   & 66.9/70.9/78.2 & 89.1/96.0/93.2 \\
        &   & K=8   & 76.0/79.5/84.1 & 94.1/98.0/96.1 \\
    \hline
    &  & K=1   & 97.0/97.6/93.9 & 94.0/98.1/96.2 \\
    \multirow{2}*{\usym{2714}}  & \multirow{2}*{\usym{2717}} & K=2   & 93.9/94.2/92.7 & 96.3/98.6/96.2 \\
        &  & K=4   & 96.2/96.4/94.4 & 96.3/98.7/96.5 \\
        &   & K=8   & 97.2/97.4/94.3 & 94.4/98.3/96.3 \\
    \hline
    & & K=1   & 68.5/72.6/77.4 & 89.9/95.7/92.1 \\
   \multirow{2}*{\usym{2717}}  & \multirow{2}*{\usym{2714}} & K=2   & 70.2/73.0/78.4 & 90.1/95.8/92.4 \\
      &    & K=4   & 84.7/88.0/82.2 & 87.6/95.2/92.9 \\
       &    & K=8   & 83.9/86.2/81.7 & 93.5/97.0/94.0 \\
    \hline
    & & K=1   & 97.6/98.3/94.7 & 94.0/98.1/96.2 \\
    \multirow{2}*{\usym{2714}} & \multirow{2}*{\usym{2714}} & K=2   & 97.4/98.2/94.6 & 96.3/98.6/96.4 \\
      &    & K=4   & 97.6/98.2/94.6 & 96.3/98.7/96.7 \\
       &    & K=8   & 97.7/98.3/94.5 & 94.5/98.0/95.4 \\
    \hline
    \end{tabular}%
    
  \label{tab:addlabel}%
\end{table}%

\begin{table}[t]
  \centering
  \caption{Ablation Studies on Different Methods of Synthesizing Anomalies. (I-AUROC/I-AUPR/F1-MAX)}
    \label{sa}
    \begin{tabular}{cccc}
    \hline
       Dataset  & Setup & \thead{Random \\ perturbation}  & NSA \\
    \hline
     & K=1   & 96.9/97.5/93.7 & 97.6/98.3/94.5 \\
    \multirow{2}*{VisA}   & K=2   & 97.2/97.9/94.5 & 97.4/98.2/94.6 \\
          & K=4   & 97.2/98.0/94.3 & 97.6/98.2/94.6 \\
           & K=8   & 96.3/97.0/93.3 & 97.7/98.3/94.5 \\
    \hline
     & K=1   & 94.0/98.1/96.2 & 94.0/97.0/95.3 \\
    \multirow{2}*{MVTEC-AD}  & K=2   & 96.3/98.6/96.4 & 95.2/98.2/96.3 \\
          & K=4   & 96.3/98.7/96.7 & 95.8/98.6/96.4 \\
           & K=8   & 94.5/98.0/95.4 & 96.1/98.6/96.0 \\
    \hline
    \end{tabular}%

  \label{sa}%
\end{table}%

\subsubsection{Comparison with Full-Shot Methods}
As illustrated in Tab.~\ref{full}, we compare our results in few-shot
setup with previous full-shot methods on VisA and MVTEC-AD datasets. Though with less normal samples, our classification performance is competitive with these full-shot methods. The state-of-the-art methods on VisA are DiffusionAD and GLASS~\cite{glass} which can reach 98.8\% I-AUROC using full-shot samples. But I-AUROC of CLIP-FSAC++ achieves 97.7\% which is very close to 98.8\% I-AUROC of DiffusionAD. CLIP-FSAC++ outperforms most full-shot methods on VisA but there continues to be a disparity compared to full-shot methods on MVTEC-AD. On MVTEC-AD dataset, we can reach I-AUROC score of 96.3\% in 2-shot and 4-shot surpassing some early relatively anomaly detection methods such as MKD~\cite{mkd} and CutPaste~\cite{cutpaste} .

\begin{figure}[t]
  \centering
  \centerline{\includegraphics[width=\linewidth]{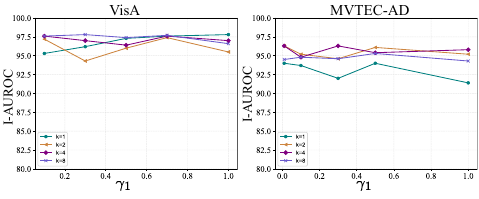}}
  \caption{Sensitiveness on coefficient of text-driven visual features.}
  \label{t}
\end{figure}

\subsection{Ablation Studies}
\subsubsection{Effect of Loss Function}
In this section, we use different loss functions to optimize CLIP-FSAC++ including cross-modality contrastive loss, cross-entropy loss and the combination of these two losses. It can be seen in the expression of loss that cross-modality contrastive loss focuses on both normal and abnormal visual and text features. It pulls the homogeneous visual and language representations and pushes the heterogeneous visual and language representations. While cross-entropy loss only concentrates on the relationship between visual features and abnormal features. Though their concerns are different, the goal of these two losses is to assign high anomaly score to abnormal samples and low anomaly score to normal samples. So we make a deep exploration in these two losses. As can be seen in Tab.~\ref{lf}, we find that anomaly classification performance is satisfactory using no matter cross-modality contrastive loss or cross-entropy loss. When only using cross-modality contrastive loss, performance on MVTEC-AD deteriorates a lot compared to CLIP-FSAC++ with 0.3\%$\downarrow$, 0.2\%$\downarrow$, 3.0\%$\downarrow$ and 0.9\%$\downarrow$ but it is improved by 0.2\% on VisA in 2-shot setup. Using cross-entropy instead of cross-modality contrastive loss hardly benefits classification performance with 0.9\%$\downarrow$, 1.6\%$\downarrow$, 0.8\%$\downarrow$, 0.2\%$\downarrow$ in VisA and 0.9\%$\uparrow$, 1.5\%$\downarrow$, 0.1\%$\downarrow$, 0\% in MVTEC-AD. When combining these two losses, CLIP-FSAC++ achieves the best results on VisA and there are a little drops on MVTEC-AD in the 1-shot and 8-shot.

\begin{figure}[t]
  \centering
  \centerline{\includegraphics[width=\linewidth]{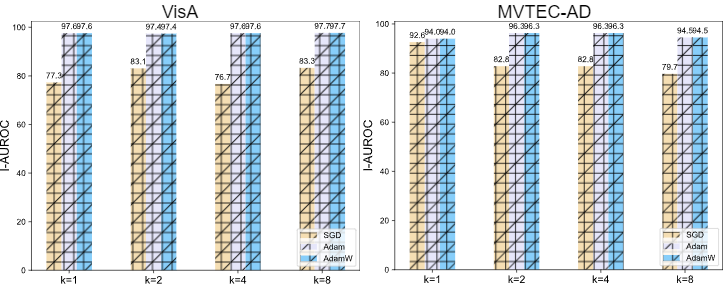}}
  \caption{Comparison of the SGD, Adam and AdamW optimizers for CLIP-FSAC++.}
  \label{opt}
\end{figure}

\subsubsection{Effect of Anomaly Descriptor}
To validate the effectiveness of anomaly descriptor in few-shot anomaly classification, we conduct a series of experiments in the light of enhancement of modality-specific driven features. The whole results are exhibited in Tab.~\ref{aad}. Firstly, we completely abandon anomaly descriptor only using two adapters. It can be seen that performance suffers from drastic drops more than 20\% on VisA. Performance on MVTEC-AD also drops to various extent. Taking a step further, we only use vision-driven text features $TF$ to enhance visual representations without text-driven visual features. Obviously, classification performance is improved a lot (72.5\% $\rightarrow$ 97.0\%,  88.3\% $\rightarrow$ 94.0\%, etc.). However, if only using text-driven visual features $VT$, the results are not improved and even deteriorate (72.5\% $\rightarrow$ 68.5\%,  89.1\% $\rightarrow$ 97.6\%, etc.) as we expected. Because if visual features are distinguished, they can be classified correctly with sub-optimal text representations. Oppositely, it is hard to match visual features which cannot be easily distinguished for text features. CLIP-FSAC++ represents a satisfactory performance with both visual and text enhancement.

\begin{figure}[t]
  \centering
 
  \centerline{\includegraphics[width=\linewidth]{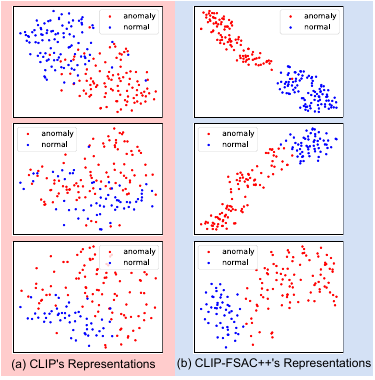}}
  \caption{T-SNE visualization of visual representations of CLIP and CLIP-FSAC++.}
   \label{scatter}
\end{figure}

\begin{figure*}
  \centering
  \centerline{\includegraphics[width=\linewidth]{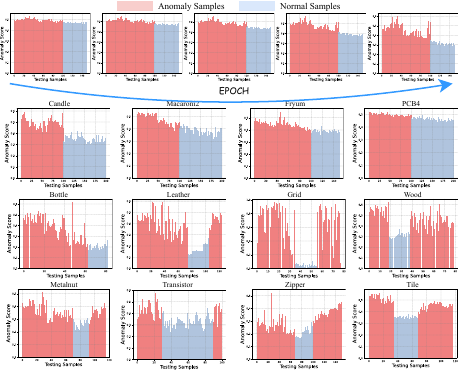}}
  \caption{Anomaly score distribution of testing samples in different
epochs and specific classes in VisA and MVTEC-AD.}
  \label{his}
\end{figure*}

\subsubsection{Effect of Synthetic Anomalies}
The way to synthesize anomalies is essential in CLIP-FSAC++ which is one of keys influencing the training results. We hope that expected synthetic anomalies are more natural and follow the realistic anomalies distribution. So we use two synthetic methods that are random perturbation and NSA. Anomalies with low quality are so misleading that our method can not classify samples correctly.
To explore the influence of synthetic anomalies and find the more natural generation method, we exchange the synthetic ways between MVTEC-AD and VisA datasets. For MVTEC-AD, we use NSA to synthesize anomalies while random perturbation is used for VisA. As shown in Tab.~\ref{sa}, the results on VisA of random perturbation are all lower than NSA which are 96.9\% (0.7\%$\downarrow$), 97.2\%(0.2\%$\downarrow$), 97.2\% (0.4\%$\downarrow$), and 96.3\%(1.4\%$\downarrow$) of I-AUROC in 1-shot, 2-shot, 4-shot and 8-shot respectively, which demonstrate that NSA can simulate the realistic anomalies in VisA dataset. On the contrary, we use NSA for MVTEC-AD dataset and there exists improvement only in 8-shot with 96.1\% (1.6\%$\uparrow$). 
Though there exists differences in anomaly classification performance with different synthetic methods, the results of our method still surpass many other previous works. The robust of CLIP-FSAC++ to synthetic anomalies is experimentally proved. 

\subsubsection{Sensitiveness on Coefficient $\gamma_1$ of Text-driven Visual Features.}
The significant improvement resulted from vision-driven text features $TF$ is showcased in Table \ref{aad}. However, it seems like that performance contribution of text-driven visual features $VT$ is subtle. Therefore, we conduct a series of ablation experiments where we change text-driven visual features coefficient $\gamma_1$ in Eq. 9. Coefficient $\gamma_1$ controls the proportion of text-driven visual features in text features. As illustrated in Fig. \ref{t}, we use five different $\gamma_1$ in 1-shot, 2-shot, 4-shot and 8-shot settings. It can be found that the performance on VisA and MVTEC-AD datasets with different $\gamma_1$ is undulating. When $\gamma_1=1$, performance on MVTEC-AD dataset in 1-shot setting drops drastically. When $\gamma_1=0.3$, performance on VisA dataset in 2-shot setting drops to 94.3\% (3.1\%$\downarrow$) of I-AUROC. It is experimentally demonstrated that appropriate coefficient $\gamma_1$ can improve performance expectantly.

\subsubsection{Different Optimizers for CLIP-FSAC++.}
Except for loss functions, optimizer is also an important factor in training stage. We use Adam, SGD and AdamW optimizers to train CLIP-FSAC++ on VisA and MCTEC-AD datasets mentioned above and compare the performance under different optimizers. It is notable that learning rates of all optimizers is 0.0005 for image adapter and 0.0001 for text adapter. Weight decay is 0 for Adam and 0.01 for AdamW. The anomaly classification results are presented in Fig. \ref{opt}. It demonstrates that the performance is the same when using Adam and AdamW optimizers. Compared with Adam and AdamW optimizers, SGD can not engender the best optimization trajectory. The performance drops dramatically when using SGD optimizer. SGD leads to 13.5\% drop and 14.8\% drop of I-AUROC in 2-shot and 8-shot on MVTEC-AD dataset. On VisA dataset, SGD accounts for more serious performance deterioration. Considering the whole performance, we use Adam optimizer to train CLIP-FSAC++.

\section{Further Analysis}

\subsection{Visualization of Feature Representation}
To further understand the adapted and enhanced visual representations, we conduct the t-SNE~\cite{tsne} visualization of test data representations. Except for adapted and enhanced visual features, we also conduct the t-SNE visualization on original visual representations from CLIP, which intuitively validates the effectiveness of our method. By t-SNE visualization method, we project high-dimension features into 2D space as shown in Fig. \ref{scatter}. The left column depicts original representations from CLIP of normal and abnormal features and the right column depicts representations from CLIP-FSAC++ of normal and abnormal features. Distribution of normal and abnormal visual features are easily observed in the low-dimension space. It is demonstrated in Fig.~\ref{scatter} (a) that there is an unexpected large overlap between normal and abnormal visual features which hinders the anomaly classification. The distribution of visual features is irregular
with adaptation and enhancement. There is no doubt that mixed visual features can not be classified correctly and easily. In stark contrast, there is a clear boundary between adapted and enhanced visual representations of normality and abnormity and the distribution of visual representations is regular as shown in the Fig.~\ref{scatter} (b). Incorporating the proposed modules in CLIP-FSAC++, normal and abnormal visual features are grouped into contracted clusters respectively with a boundary.

\subsection{Distribution of Anomaly Scores}
In this section, we visualize the anomaly score distribution of testing samples depicted by histograms. The first row of Fig.~\ref{his} shows the anomaly score distribution of exampled class "cashew" in VisA in different epochs. It can be found that the gap between normal samples and anomaly samples becomes larger as epoch increases. Ideally, anomaly score of anomaly samples are all higher than normal samples, which means that we can easily classify them according to anomaly score. The anomaly score distribution of specific classes in VisA and MVTEC-AD such as "Candle", "Fryum", "Grid", "Wood", etc. As can be seen from Fig.~\ref{his}, anomaly score distribution of testing samples is satisfactory as anomaly score of anomaly samples are generally larger than normal samples. Especially in anomaly score distribution of class "Grid", we can find that the anomaly score gap is dramatically considerable and anomaly score of anomaly samples in "Grid" are all higher than normal samples. So we experimentally figure out why I-AUROC of class "Grid" reaches 100\% shown in Tab \ref{iaurocmvtec}. However, there exists inverted situations in anomaly score due to high degree of similarity between anomaly and normal samples.
Fig.~\ref{his} demonstrates that our CLIP-FSAC++ is effective and adapts CLIP for better performance in few-shot anomaly classification task.

\begin{figure}
\centering
\includegraphics[width=3.25in, keepaspectratio]{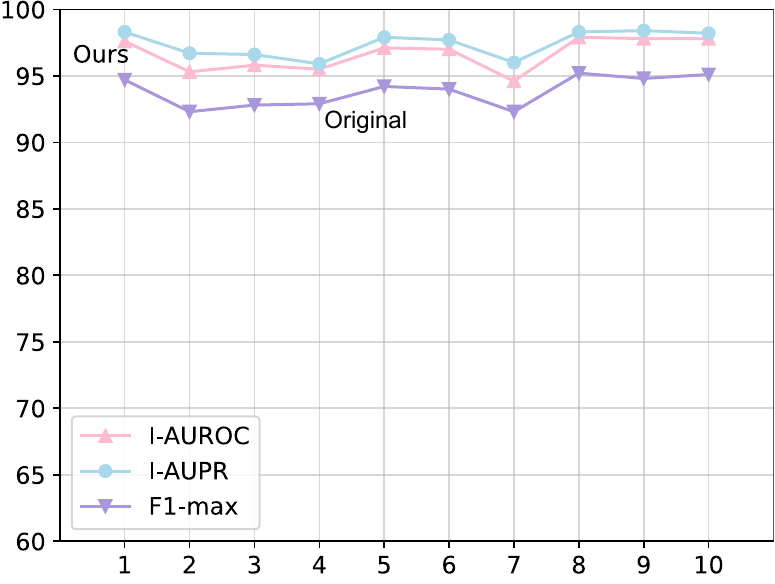}\\
\caption{Performance comparison of different text prompts. Ours is
compositional prompt ensemble used in CLIP-FSAC++ and Original is the simplest text prompts proposed in CLIP.}
\label{prompts}
\end{figure}

\subsection{Performance Comparison of Different Text Prompts.}
Text prompts are particularly vital in vision-language models. The design of text prompts is deeply explored no matter manually crafting prompts or learnable prompts. So we take a closer look at influence of text prompts in CLIP-FSAC++. We try 10 different text prompts in this experiment. The results with different text prompts are shown in Fig. \ref{prompts}. $Ours$ means our used compositional prompt ensemble in CLIP-FSAC++ and $Initial$ means initially proposed text prompts in CLIP, that is, \textsf{A photo of a [cls]} and \textsf{A photo of a damaged [cls]}. It is demonstrated that anomaly classification performance fluctuates softly though we change text prompts. 
As shown in Fig. \ref{prompts}, I-AUROC of CLIP-FSAC++ is always larger than 95\% when using different text prompts, surpassing all other state-of-the-art few-shot anomaly detection methods. Even if using the simplest text prompts, performance of CLIP-FSAC++ is still satisfactory. Different from learnable prompts \cite{coop}, we optimize representations of language to better match relative visual features. CLIP-FSAC++ not only demonstrates robustness against synthetic anomalies but also shows robustness to text prompts.

\section{Why our proposed Anomaly Descriptor is effective?}
\begin{figure}
\centering
\includegraphics[width=3.33in, keepaspectratio]{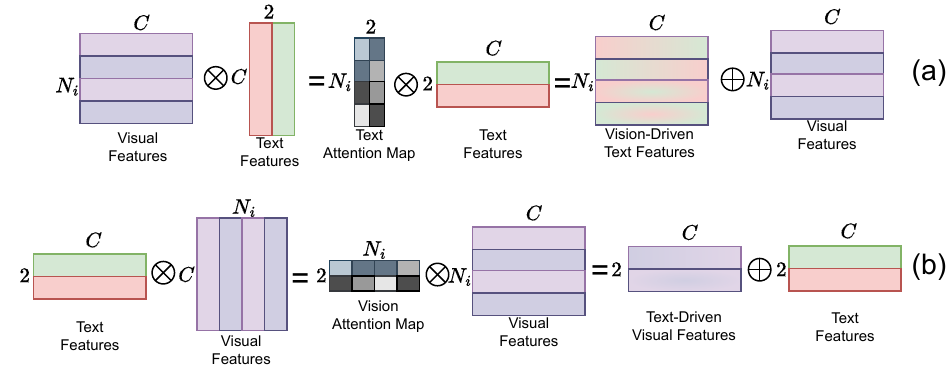}\\
\caption{The attention calculation in our proposed anomaly descriptor.}
\label{adc}
\end{figure}

In this section, we delve into our proposed Anomaly Descriptor. It is demonstrated in Table \ref{aad} that text-driven visual features and vision-driven text features from Anomaly Descriptor improve anomaly classification performance a lot, especially vision-driven text features. We analyze the reason of performance improvement caused by these modality-specific driven features. We firstly dissect the calculation of vision-driven text features and how they influence the visual representations. As shown in Fig. \ref{adc} (a), we calculate the text attention weights between visual features and text features. Text attention weight presents the similarity between each visual features and text features. Based on text attention weight, we calculate weighted sum of text features for each visual feature to attain vision-driven text features. Afterward, vision-driven text features are added to visual features. 
We inject the text information into visual representations as language priors. In final stage, the matching ability between visual and text embeddings is facilitated by modality-specific priors. In other words, the weight of normal text feature is larger than abnormal text feature for normal visual features. So more normal text information is injected into normal visual features, leading to large similarity between normal visual features and text features in final stage. So were abnormal visual and text features. Similarly, text-driven visual features is weighted sum of visual features based on similarity between visual and text features and then they are injected into text features as visual priors.

\section{Conclusion}
In this paper, we propose a novel fine-tuning framework named CLIP-FSAC++ to adapt CLIP for few-shot anomaly classification. We use two adapters to embed image and text features into a tight space for better alignment. Meanwhile, we propose a a cross-modality interaction module named anomaly descriptor to enhance the correlation between vision and language consisting of two cross-modality attention modules. The adapters and anomaly descriptor are optimized jointly. Our method achieves state-of-the-art results on VisA and MVTEC-AD datasets, even outperforming many full-shot methods and the effectiveness and robust of proposed CLIP-FSAC++ are demonstrated by ample experiments and visualization. 

We believe that cross-modal description ability of the vision-language pre-training model still has a lot of room for improvement in anomaly classification and use of CLIP is costly in real-world scenarios. In the future, we will focus on hard classes mentioned above and improve the matching ability of CLIP-FSAC++ to classify these inconspicuous anomaly samples correctly and adapt strong representative ability and generalization of CLIP to real-world scenarios in a light way.

\bibliographystyle{IEEEtran}
\bibliography{IEEEabrv,reference}

\end{document}